\documentclass{IEEEtran}
\usepackage{hyperref}
\hypersetup{breaklinks=true}
\usepackage[strings]{underscore}
\usepackage{cite}

\ifCLASSINFOpdf
  \usepackage[pdftex]{graphicx}
  \graphicspath{{../figs/}}
  \DeclareGraphicsExtensions{.pdf,.jpeg,.png}
\else
\fi
\begin{document}
%
\title{Autonomous Vehicles: Open-Source Technologies, Considerations, and Development}
%
%
%
\author{Oussama~Saoudi,
        Ishwar~Singh,
        and~Hamidreza~Mahyar
\thanks{O. Saoudi was with the Department of Computing and Software, McMaster University, Canada}
\thanks{I. Singh and H. Mahyar are with W Booth School of Engineering Practice and Technology, Faculty of Engineering, McMaster University, Canada (e-mail: mahyarh@mcmaster.ca)}
\thanks{Manuscript received September, 2021; revised September, 2021.}}

%
%

\markboth{Preprint}%
{Saoudi \MakeLowercase{\textit{et al.}}: Autonomous Vehicles: Open-Source Technologies,Considerations, and Development}
%



\maketitle

\begin{abstract}
Autonomous vehicles are the culmination of advances in many areas such as sensor technologies, artificial intelligence (AI), networking, and more. This paper will introduce the reader to the technologies that build autonomous vehicles. It will focus on open-source tools and libraries for autonomous vehicle development, making it cheaper and easier for developers and researchers to participate in the field. The topics covered are as follows. First, we will discuss the sensors used in autonomous vehicles and summarize their performance in different environments, costs, and unique features. Then we will cover Simultaneous Localization and Mapping (SLAM) and algorithms for each modality. Third, we will review popular open-source driving simulators, a cost-effective way to train machine learning models and test vehicle software performance. We will then highlight embedded operating systems and the security and development considerations when choosing one. After that, we will discuss Vehicle-to-Vehicle (V2V) and Internet-of-Vehicle (IoV) communication, which are areas that fuse networking technologies with autonomous vehicles to extend their functionality. We will then review the five levels of vehicle automation, commercial and open-source Advanced Driving Assistance Systems, and their features. Finally, we will touch on the major manufacturing and software companies involved in the field, their investments, and their partnerships. These topics will give the reader an understanding of the industry, its technologies, active research, and the tools available for developers to build autonomous vehicles.
\end{abstract}


%
\IEEEpeerreviewmaketitle

\title{Autonomous Vehicles: Open-Source Technologies, Considerations, and Development}

\author{\name Oussama Saoudi \email saoudio@mcmaster.ca \\
       \addr  Faculty of Engineering\\
       McMaster University\\
       Hamilton, Ontario, Canada
       \AND
       \name Ishwar Singh \email isingh@mcmaster.ca \\
       \addr  Faculty of Engineering\\
       McMaster University\\
       Hamilton, Ontario, Canada
       \AND
       \name Hamidreza Mahyar \email mahyarh@mcmaster.ca \\
      \addr Faculty of Engineering \\
       McMaster University\\
       Hamilton, Ontario, Canada}

\maketitle

\begin{abstract}
Autonomous vehicles are the culmination of advances in many areas such as sensor technologies, artificial intelligence (AI), networking, and more. This paper will introduce the reader to the technologies that build autonomous vehicles. It will focus on open-source tools and libraries for autonomous vehicle development, making it cheaper and easier for developers and researchers to participate in the field. The topics covered are as follows. First, we will discuss the sensors used in autonomous vehicles and summarize their performance in different environments, costs, and unique features. Then we will cover Simultaneous Localization and Mapping (SLAM) and algorithms for each modality. Third, we will review popular open-source driving simulators, a cost-effective way to train machine learning models and test vehicle software performance. We will then highlight embedded operating systems and the security and development considerations when choosing one. After that, we will discuss Vehicle-to-Vehicle (V2V) and Internet-of-Vehicle (IoV) communication, which are areas that fuse networking technologies with autonomous vehicles to extend their functionality. We will then review the five levels of vehicle automation, commercial and open-source Advanced Driving Assistance Systems, and their features. Finally, we will touch on the major manufacturing and software companies involved in the field, their investments, and their partnerships. These topics will give the reader an understanding of the industry, its technologies, active research, and the tools available for developers to build autonomous vehicles.
\end{abstract}


\section{Introduction}
%
%
%
%
A vehicles have been a widely researched topic that is witnessing rapid development in recent years. This paper will review the state-of-the-art technology in autonomous vehicles, focusing on free, open-source software. Open-source development has the advantage of minimizing costs for developers breaking into the field and accelerating research done in the field by sharing preexisting work. Many fields of research and software engineering intersect to create autonomous vehicles. The topics are chosen to give the reader a broad view of the essential components that make up autonomous vehicles such as the sensors, embedded operating system, and autonomous driving assistance system. We will also discuss supplementary topics that enhance autonomous vehicles and their development such as vehicle-to-vehicle communication and vehicle simulators.

Sensors are a core component of Autonomous vehicles and their advantages and disadvantages have been widely discussed \cite{SensorReview,AVSensorReview,RadarRange,FutureRadar,RadarReview,LiDARReview}. Specific surveys have also been compiled for radar \cite{RadarRange,FutureRadar,RadarReview} and LiDAR \cite{LiDARReview}.
Several studies have been published discussing certain technologies that go into autonomous vehicles. SLAM is among the highly studied fields with papers surveying the algorithms, classifications, and theory \cite{SLAMDef,SLAMHistory,SLAMDatasets,SLAMFuture,c9,c8,DLSLAMSurvey,ClassicalSLAM}.
Surveys have also been conducted on IoV technologies \cite{V2I,5G_CV2X,c74,V2V,c81,V2XTech,V2P,DSRCCV2V}.
While these surveys cover a significant depth in each of the fields, there are no broad reviews of autonomous vehicle technologies that cover the mentioned fields holistically. 

We will now briefly outline the sections in the paper. Section~\ref{sec:Sensors} will discuss different sensors used in autonomous vehicles, their benefits, and constraints. Section~\ref{sec:SLAM} will cover Simultaneous Localization and Mapping (SLAM), a method of mapping a vehicle's environment that allows it to navigate within it. Section~\ref{sec:Simulator} will discuss open-source simulators used to test and train deep learning models for robust autonomous vehicles at a low cost. Section~\ref{sec:EmbeddedOS} will discuss embedded operating systems used in autonomous vehicles and the considerations when choosing one. Section~\ref{sec:V2X} discusses Internet of Vehicle (IoV) and Vehicle-to-Everything (V2X) technology and its underlying network technologies. The final two sections will focus on the current commercial market of autonomous vehicles. Section~\ref{sec:ADAS} will introduce the five levels of vehicle automation, the commercially available Advanced Driving Assistance Systems (ADAS), and the features they offer. Finally, Section~\ref{sec:Commercial} will look at the landscape of major manufacturing and software companies and the progress, investments, and partnerships they have made to bring autonomous vehicles to the market. This paper will leave the reader with an understanding of the unique complexities and challenges of autonomous vehicles and the core technologies that compose them.

\section{Sensors}
\label{sec:Sensors}
A critical component to consider when designing an autonomous vehicle is the choice of sensors. Each one has its advantages and disadvantages, determines the type of algorithm used, and impacts the conditions in which the vehicle can operate. Autonomous vehicles may use multiple sensor modalities, and some sensors are better suited to certain constraints and environmental conditions than others. Another consideration is the cost; developers and manufacturers may favor one sensor over others to minimize costs.  We will discuss the following sensors: LiDAR, Radar, Camera, and Ultrasonic.
    \subsection{LiDAR}
\begin{figure}[h]
        \centering
        \includegraphics[width=0.4\textwidth]{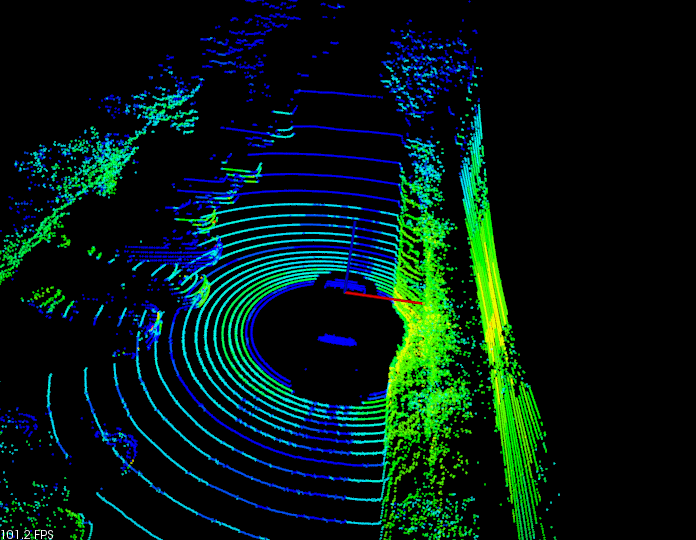}
        \caption{Example of point cloud generated by Velodyne LiDAR \cite{LidarIMG}}
        \label{fig:LidarFig}
    \end{figure}

    LiDAR stands for Light Detection and Ranging and is a popular sensor for use in autonomous vehicles. It outputs point clouds which are 3D representations of surfaces in its environment. These point clouds have a resolution at around the centimeter level. LiDAR is a long-range sensor, providing maximum ranges over 250m \cite{SensorReview}. It utilizes TOF (Time of Flight) to measure the distances of objects to the sensor. In addition, some types of LiDAR can accurately detect the velocity of moving objects \cite{LiDARReview}. One of the primary disadvantages of LiDAR is its steep costs, with models typically ranging from \$4000-\$10,000. However, technological advancements such as solid-state LiDAR promise to bring down the prices dramatically in the coming years \cite{AVMarketRsrch}. The second drawback to LiDAR is its poor performance in harsh weather conditions. Fog, snow, and rain all significantly reduce the quality and range of detection due to the light scattering effect of the water droplets \cite{LiDARReview}. Despite the drawbacks, LiDAR is one of the most popular sensors due to its excellent point cloud resolution and range. Fig. \ref{fig:LidarFig} shows an example of point cloud generated by Velodyne LiDAR.

    \subsection{Radar}
    Radar uses radio waves to find the distance of objects to the sensor. As with LiDAR, radar forms 3D point clouds to represent its surroundings digitally, and it uses TOF to measure the distances to surfaces. Radar can sense in several ranges and is categorized into ultra-short-range, short-range, and long-range with the maximum range going up to 250m \cite{RadarRange}. A unique advantage to radar is the transvision effect, which utilizes reflections of radio waves to see through obstacles and detect other objects out of the vehicle's direct line of sight. It allows a radar sensor to gather more information about its surroundings \cite{RadarReview}. Radar has the advantage of being resilient to environmental conditions such as high luminosity, rain, fog, snow, and dust. In addition, radar is a far cheaper sensor than LiDAR. A drawback of radar sensors is the effects of radio wave interference on the sensor. The source of the interference is threefold: Interference from the sensor on itself, interference from other radar sensors on the vehicle, and interference from sensors on other vehicles \cite{FutureRadar}. This limitation is significant because dense traffic and widespread adoption of autonomous vehicles equipped with radar may worsen the interference. Radar sensors' cheap cost and resilience to environmental conditions make them an attractive choice for manufacturers and developers of autonomous vehicles. 

    \subsection{Camera}
    Cameras are vital sensors for autonomous vehicles since they are the only sensor the provides RGB data. They capture information such as colors and texture, which allows the vehicle to detect road signs, traffic lights, and other objects. There are two types of cameras. Regular cameras can capture visual information. Some processing algorithms on stereo cameras can detect depth using only vision. The second type of camera is the RGB-D camera. It provides color image and depth estimates for each pixel by using both a camera and a projector. The projector emits a pattern of light, which allows the sensor to determine the depth of each point based on the camera view of the pattern \cite{RGBDCam}. RGB-D cameras have become a mainstay in SLAM. An example of an RGB-D camera that can be used for autonomous vehicles  is the OAK-D camera developed by OpenAI \cite{OAKD}. Another advantage of cameras is their wide availability and cheap cost, making them easy to integrate into a vehicle. The main limitation of cameras is their susceptibility to adverse weather conditions such as rain, snow, fog, and low luminosity. They are the sensor most inhibited by weather conditions \cite{SensorReview}. Thermographic cameras can mitigate the adverse effects of low luminosity by detecting infrared light \cite{AVSensorReview}. Cameras are cost-effective and are essential for detecting visual information such as road signs and traffic lights.
    
    \subsection{Ultrasonic}
    Ultrasonic sensors utilize sound waves and TOF to measure the distance to objects. They are the most accurate sensor for close range. It has the advantage of being resilient to adverse weather conditions like rain or snow. It is also the cheapest sensor available for autonomous vehicles. Ultrasonic sensors are already widely used as parking sensors to assist drivers in gauging their distance to an obstacle when parking \cite{AVSensorReview}. Ultrasonic sensors have proven helpful in previous vehicle applications and may be an asset when requiring short-range sensor data.

\section{Simultaneous Localization and Mapping}
\label{sec:SLAM}
Simultaneous Localization and Mapping has been a hotly researched field in robotics for many years \cite{SLAMHistory}. It involves utilizing sensor data to perform two functions in a robot: The first is to construct a 2D or 3D digital model of the robot's environment using sensor data, which is known as mapping; The second function of SLAM is to determine the robot's state and position in the environment it has mapped, i.e., localize the robot \cite{SLAMFuture} \cite{SLAMDef}. SLAM is a vital element in autonomous vehicles due to its utility in environment perception, which aids in intelligent response such as avoiding obstacles, path planning, and maintaining a consistent mapping from a virtual to a real-world position. The types of SLAM algorithms can be divided into classical and deep learning-based methods \cite{DLSLAMSurvey}. \cite{ClassicalSLAM} details the categories and methods used in classical SLAM algorithms, such as filters and graph-based methods. Deep learning algorithms can either perform the entire SLAM process or be part of a SLAM pipeline. Many existing libraries implement classical SLAM algorithms, but deep learning models may become more widely adopted. SLAM methods may leverage different sensor modalities \cite{SLAMHistory} such as cameras, LiDAR, and radar. There also exist several datasets and benchmarks which compare the performance of the SLAM algorithms \cite{SLAMDatasets} \cite{OpenDatasets}. One of the dominant datasets used for SLAM algorithm benchmarking has been the KITTI dataset\cite{KITTI}, which offers two sensor modalities: stereo camera and LiDAR. There are many new exciting datasets being released that offer different sensor modalities, locations, and dataset sizes such as Argoverse \cite{Argoverse} and Argoverse 2 \cite{wilson2021argoverse}, Lyft L5 \cite{c71}, Waymo Open\cite{c70}, and nuScenes\cite{c64}. Open-source libraries make implementing the SLAM algorithms into code easier and bug-free; thus, we will mention libraries available for each category of SLAM. The next sections will cover classical Visual, LDIAR, and Radar SLAM, as well as deep learning-based methods. We also list notable algorithms in Table \ref{table:SLAM}.

\label{table:SLAM}
    \begin{table*}[t]
        \resizebox{\linewidth }{!} {
            \begin{tabular}{ | l ||  p{0.65\linewidth}| }
                \hline
                Type of SLAM & Notable Examples\\
                \hline
                Visual SLAM (Monocular)& Large Scale Direct Monocular SLAM (LSD-SLAM) \cite{c12}, ORB-SLAM \cite{c13}, and Open Structure from Motion (OpenSfM) \cite{c14}\\
                \hline
                Visual SLAM (Stereo, RGB-D) &ORB-SLAM2 \cite{c16}, OpenVSLAM\cite{c17}, UcoSLAM\cite{c18}, and GSLAM \cite{c19}.\\
                \hline
                LiDAR SLAM & LiDAR Odometry and Mapping (LOAM) algorithm \cite{c20},Fast-LOAM \cite{c21}, Lightweight and Ground Optimized LOAM (LeGO-LOAM) \cite{c22}, Intensity Scan-Context LOAM (ISCLOAM) \cite{c23}, LOAM-Velodyne \cite{c24}, and LOAM-Livox \cite{c25}, LIO-SLAM \cite{c26}, HDL-graph \cite{c20.5}, MULLS \cite{c27}, and Google Cartographer \cite{c28}\\
                \hline
                Radar SLAM & RadarSLAM \cite{c29}, Landmark-based Radar SLAM \cite{LandmarkRadar}, Radar Scan Matching using Fourier-Mellin Transform \cite{RadarScanMatching}, and Real-time Pose Graph Radar SLAM \cite{RadarPoseGraph}\\
                \hline
            \end{tabular} 
        }\\
        \caption{SLAM Algorithms used for each type of Sensor}
    \end{table*}

    \subsection{Visual SLAM}
    Visual SLAM algorithms only use cameras to perform SLAM. It is an attractive option to minimize the cost and the sensor modalities needed on the vehicle. Some of the latest visual algorithms have competed with and even outperformed ones that use LiDAR on KITTI. These algorithms may work on one or more of the following camera types: monocular, stereo, and RGB-D. Visual-inertial SLAM is a class of multimodal visual SLAM algorithms that uses an inertial measurement unit (IMU) along with one of the aforementioned camera types. The coupling of IMU with cameras helps with scale ambiguity and robustness to motion blur. \cite{c8}
    We will now discuss available packages, frameworks, and libraries for visual SLAM and visual-inertial SLAM. 

    \medskip

    Visual-inertial SLAM is used to reduce the location drift within the map, typically increasing the accuracy of visual methods.

    Visual and visual-inertial SLAM continues to develop, has many existing solutions, and has shown great promise in its performance despite only using cameras. 
    
    \subsection{LiDAR SLAM}
    LiDAR is another widely used sensor in SLAM. It produces a point-cloud output representing the environment and is one of the two sensors in the KITTI dataset and benchmark. LiDAR is the sensor of choice for many top-performing algorithms in the KITTI performance rankings. Many algorithms are based upon the LiDAR Odometry and Mapping (LOAM) algorithm \cite{c20} due to its high performance on the KITTI dataset. Many LiDAR SLAM algorithms are derived from the LOAM algorithm \cite{c21, c22, c23, c24, c25}. There are other LiDAR SLAM algorithms not derived from LOAM as well \cite{c26, c20.5, c27}. Google Cartographer is a 2D SLAM framework \cite{c28} which differentiates it from the other 3D SLAM packages. These are all powerful SLAM algorithms for use in autonomous vehicle development.

    \subsection{Radar SLAM}
    Radar sensors are uncommonly used in SLAM but have been investigated as a weather-robust SLAM technique. Weather conditions such as rain, fog, and snow may be detrimental to the performance of cameras and LiDAR, unlike radar. 
    Radar SLAM is a relatively unexplored but promising field and can be an asset to autonomous vehicle designs that demand more robustness. 

    \subsection{Deep Learning SLAM}
    Deep learning SLAM methods have shown great promise in improving upon classical counterparts for visual, LiDAR, radar, and fusion SLAM. \cite{DLSlam} is an overview of learning-based SLAM and suggests that artificial intelligence (AI) and deep learning models can aid performance in cases with imperfect sensor measurement, environmental dynamics, or noise. Another application of deep learning in SLAM has been the improvement of sensor fusion in SLAM and end-to-end autonomous driving \cite{MultiModalE2ESLAM, MultiModalDLSLAM, CrossModalSLAM}. 

\section{Driving Simulators}
\label{sec:Simulator}
  \begin{table*}[t]
        \resizebox{\textwidth }{!} {
            \begin{tabular}{ | l || p{0.20\linewidth} | p{0.15\linewidth} | p{0.25\linewidth} |  }
                \hline
                Simulator& Sensors & Environment Features & Other Features\\
                \hline
                CARLA           & Depth camera, RGB camera, optical flow camera, semantic segmentation camera, DVS, collision detector, lane invasion detector, obstacle detector, gnss, IMU, radar, semantic LiDAR &Lighting, fog, cloudiness, precipitation, ambient occlusion& Roadrunner map customization, traffic and pedestrian behaviour\\
                \hline
                SUMMIT          & Depth camera, RGB camera, optical flow camera, semantic segmentation camera, DVS, collision detector, lane invasion detector, obstacle detector, gnss, IMU, radar, semantic LiDAR &Lighting, fog, cloudiness, precipitation, ambient occlusion& Roadrunner map customization, traffic and pedestrian behaviour, aggression, traffic attentiveness. Has motorcycle, bus, bicycle assets\\
                \hline
                SVL             &   GPS, IMU, radar, RGB camera, segmentation sensor &Rain, wetness, fog, cloudiness, time of day, road damage& Detailed map annotation, customize road agents' spawn, path, and movement\\
                \hline
                PGDrive         & LiDAR, depth sensor, RGB Camera &None & Procedural generation, NPC traffic\\
                \hline
                Microsoft Airsim&LiDAR, RGB Camera &Time of day& None\\
                \hline
            \end{tabular}
        }\\
        \caption{A table summarizing features and sensors available in each simulator}
    \end{table*}
    
Autonomous vehicles use machine learning to perform motion planning. An immense amount of data is required to train the machine learning algorithms for motion planning and decision. Simulators offer a cheap and safe way to train machine learning algorithms for billions of miles and test the performance of autonomous vehicle software. Thus, testing and training autonomous vehicles can be made faster, cheaper, and safer with simulation. Companies like Google and Baidu drive millions of miles in simulation to collect data to train their autonomous vehicle models \cite{CBInsights}. The cost of driving millions of real-world miles is significantly greater than a comparable distance in simulation; therefore, simulation is a vital asset to developers. Simulators allow testers to observe the performance of their autonomous vehicle model in various weather conditions and road types to produce robust software that can adapt to any situation. Safety is a key consideration when deploying autonomous vehicles in the real world. Simulators eliminate safety issues, and there is no liability or danger in training in simulation. They also allow for testing responses to different agents such as cars, motorcycles, bicycles, and pedestrians. These agents are designed to model real-world behavior so that vehicles can appropriately react in different scenarios. Training with various obstacles will help the vehicle become robust to more obstacle scenarios, thus ensuring the passengers' and vehicles' safety. \cite{SimulatorOverview} outlines proprietary and open-source simulators, categorizes, and evaluates them for quality of implementation. We will only cover the open-source simulators. The categories we will focus on are graphics, vehicle dynamics, and traffic simulators. Graphics and dynamics simulators commonly use existing simulation engines to display realistic graphics. Realistic graphics allow for visual methods to be trained on data that more closely resembles its real-life usage. Two popular engines are Unreal Engine \cite{UnRealEng} and Unity \cite{Unity}, which help to create a realistic simulation of the environment, weather, and agents on the road. In this paper, the simulators we discuss are CARLA \cite{c31}, SUMMIT \cite{c32}, SVL \cite{c37}, PGDrive \cite{c34}, Microsoft Airsim \cite{c33} for simulation of vehicle dynamics and graphics, and SUMO \cite{c35}  and Flow \cite{c36} for traffic simulation. Table I summarizes features and sensors available in each simulator.

    \subsection{CARLA}
\begin{figure}[h]
        \centering
        \includegraphics[width=0.45\textwidth]{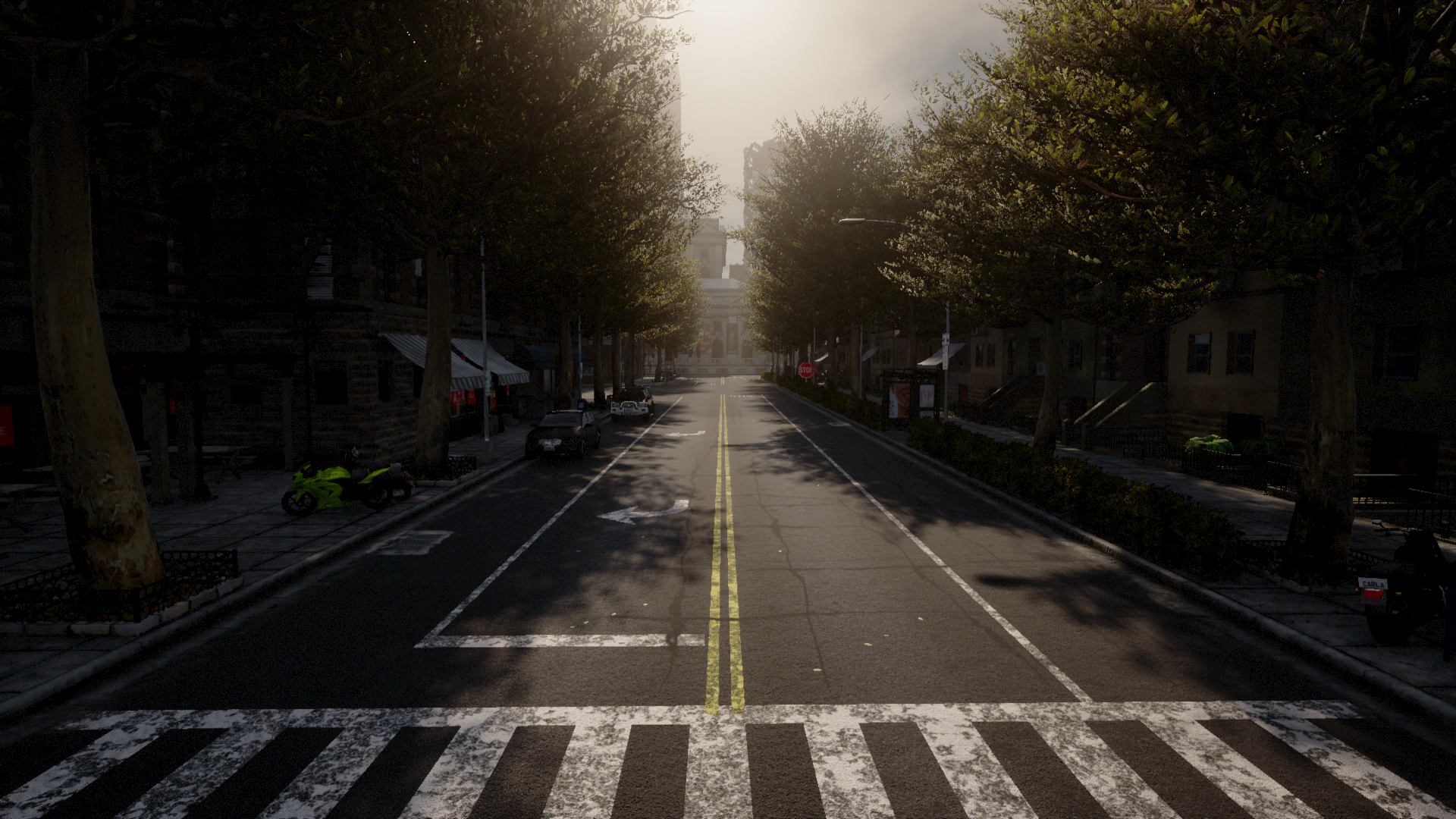}
        \caption{Screenshot of CARLA graphics in simulation \cite{CARLAIMG}}
        \label{fig:CARLAFig}
    \end{figure}

    CARLA \cite{c31} is an autonomous vehicle simulation platform with a rich set of features to train or test an autonomous vehicle, built on Unreal Engine. CARLA has models of real-world exteroceptive, interoceptive sensors, and ground truth sensors. The exteroceptive sensors include Radar, LiDAR, RGB Camera, and Dynamic Vision Sensor (DVS). The interoceptive sensors include GNSS and IMU sensors. Finally, the ground truth sensors include semantic segmentation camera, semantic LiDAR, and lane invasion detection. The user can adjust weather and environmental parameters, making their software more robust to different weather conditions. CARLA offers nine weather conditions and two lighting conditions. The nine weather conditions differ in fog, cloudiness, precipitation, and ambient occlusion, among others. The two lighting conditions include midday and sunset. The density of cars and pedestrians and the behavior of non-player vehicles can also be modified. Another helpful feature is map customization with Roadrunner, which facilitates the creation of custom maps. The primary limitations of CARLA are the small asset library and premade maps, which can lead to overfitting deep learning models to the map or assets present in the simulator. Fig. \ref{fig:CARLAFig} is a screenshot of CARLA graphics in simulation environment.

    \subsection{SUMMIT}
    SUMMIT \cite{c32} is built on top of CARLA, and so it provides the same environment and sensor features. Where SUMMIT differs from CARLA in the behavior of agents and traffic in the simulation. Agents have more adjustable parameters which dictate their behavior, giving more cases to train a model. These features include aggression and traffic attentiveness. SUMMIT provides a broader class of agents on the street to include motorcycles, bicycles, and buses. All NPC agents are controlled by an algorithm to model real-world heterogeneous crowds.  SUMMIT can generate simulations from real-world maps using OpenStreetMap, thus extending the map-making functionality of CARLA.

    \subsection{SVL}
\begin{figure}[h]
        \centering
        \includegraphics[width=0.4\textwidth]{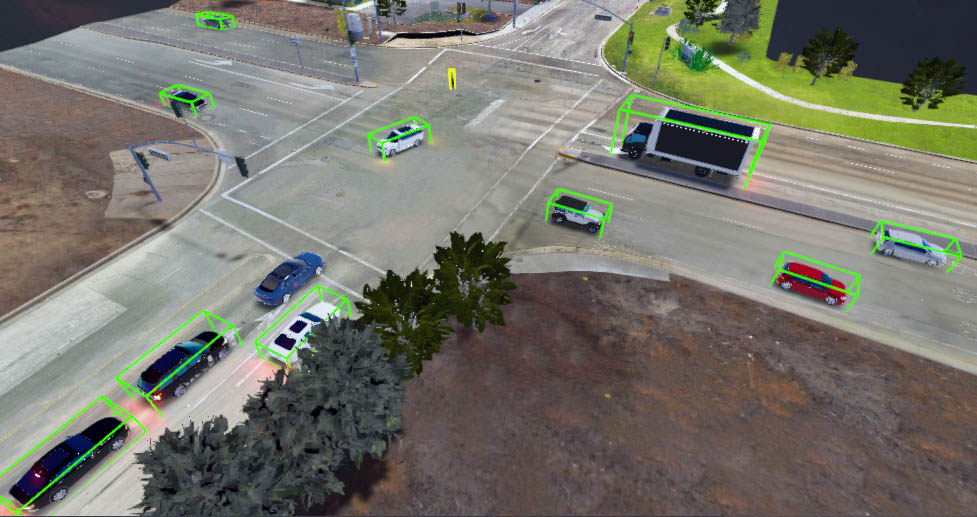}
        \caption{Screenshot of SVL graphics in simulation \cite{SVLIMG}}
        \label{fig:SVLFig}
    \end{figure}

    SVL \cite{c37} is a simulator developed by LG built on the Unity Engine. It offers many variables that can be modified to manufacture scenarios. The available exteroceptive sensors are LiDAR, radar, and camera. However, plugins allow the addition of new sensors. SVL provides ground truth sensors and real-world sensor models to simulate specific components instead of generic sensors. The user can control several environmental variables such as time of day, weather, and road condition. Weather conditions include rain, wetness, fog, and cloudiness. The road's damage is also an adjustable variable. Agents such as cars and pedestrians can be controlled with a script to adjust their spawn, paths, their movement distribution. The user can create maps out of existing 3D environments. Map annotations provide detailed information about the map, such as traffic lanes, lane boundaries, traffic signs and signals, and pedestrian walking routes. SVL also makes use of SCENIC \cite{c37.5}, which is a formal language to define testing scenarios. With SVL, the user can train an autonomous vehicle for many different scenarios, and it proves to be a powerful simulator with these options and features. Fig. \ref{fig:SVLFig} shows a screenshot of SVL graphics in a simulation environment.

    \subsection{Microsoft Airsim}
\begin{figure}[h]
        \centering
        \includegraphics[width=0.4\textwidth]{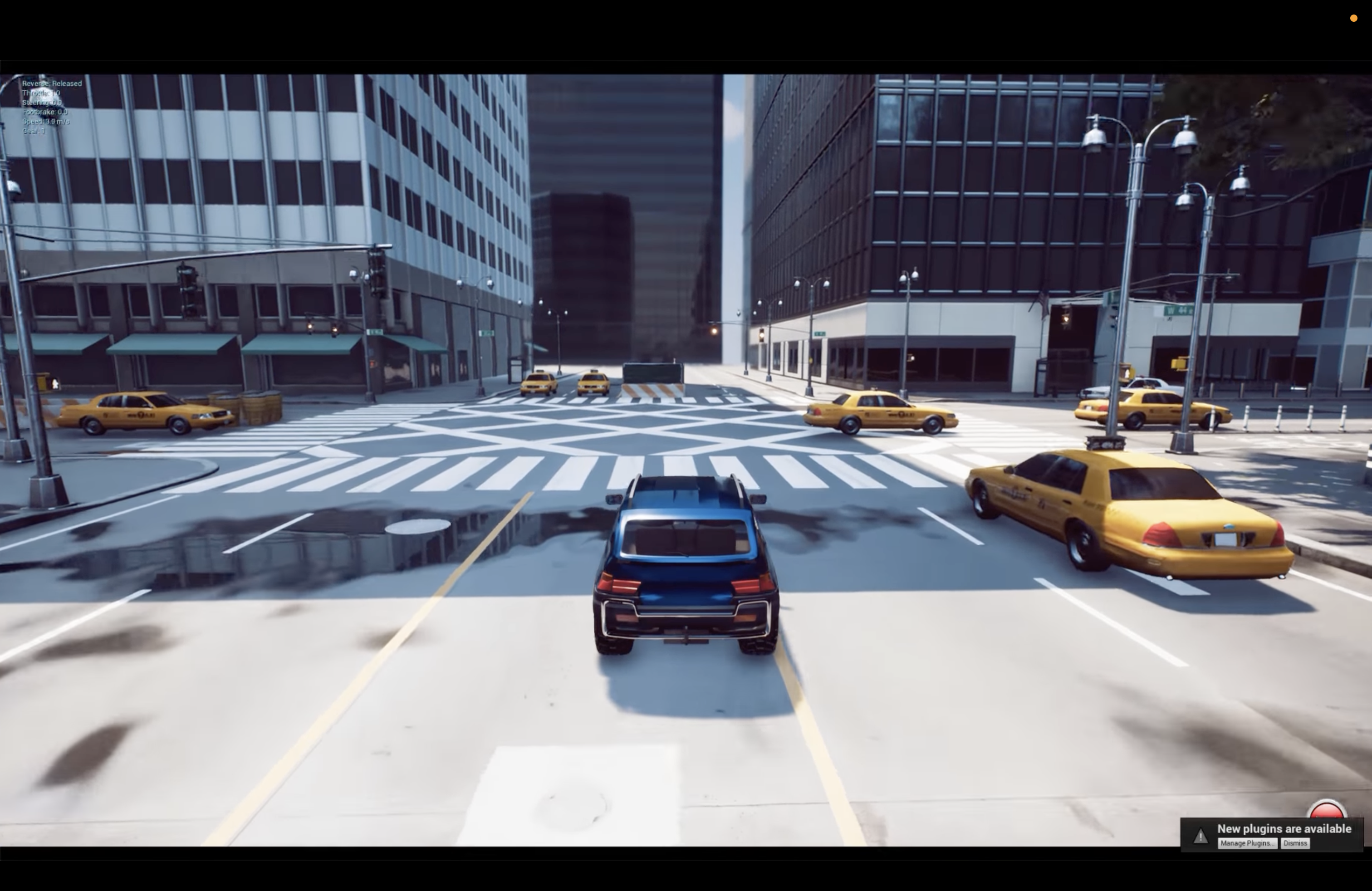}
        \caption{Screenshot of Airsim graphics in simulation \cite{AirsimIMG}}
        \label{fig:AirSim}
    \end{figure}

    Microsoft Airsim \cite {c33} is a robotics simulator that can simulate both autonomous vehicles and drones. Airsim provides camera and LiDAR sensors as well as ground truth distance sensors. It provides IMU and GPS as the offering of interoceptive sensors. Unlike CARLA, the only configurable environmental setting is the time of day. Another limitation is the lack of NPC vehicles or pedestrians, which is a significant component of training a safe and reliable autonomous vehicle. Fig. \ref{fig:AirSim} depicts a screenshot of Airsim graphics in simulation.

    \subsection{PGDrive}
\begin{figure}[h]
        \centering
        \includegraphics[width=0.4\textwidth]{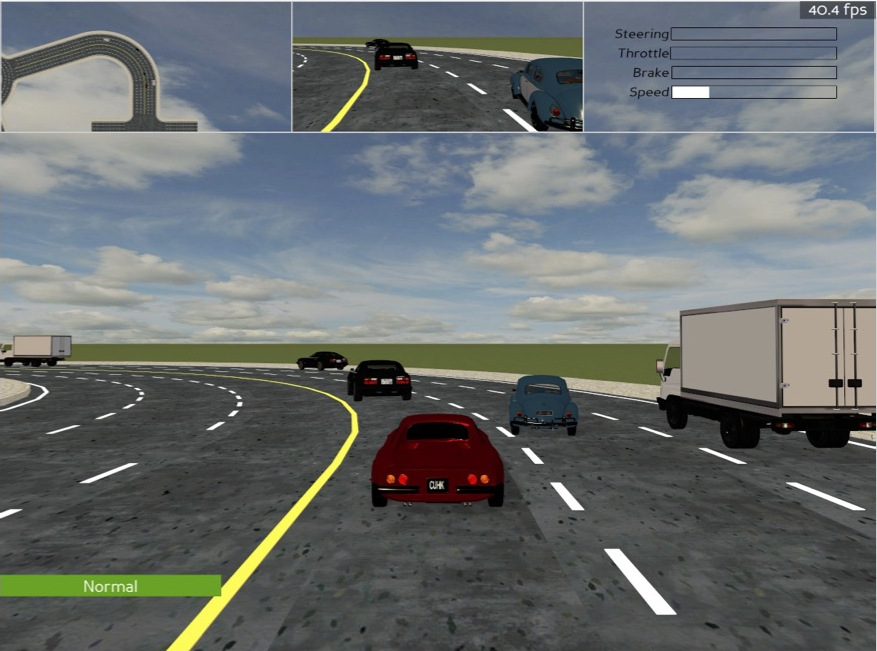}
        \caption{Screenshot of PGDrive map and graphics in simulation \cite{PGDriveIMG}}
        \label{fig:PGDriveFig}
    \end{figure}

    PGDrive \cite{c34} is a unique simulator that focuses on the use of procedural generation to create its maps. This simulator is sparse in its sensor options, only providing an RGB camera and LiDAR. It also lacks sophisticated environmental factors such as weather and lighting conditions. PGDrive does, however, provide the option for NPC traffic. Where PGDrive truly shines is in its procedural generation of maps. This functionality builds an infinite number of roads from a set of elementary blocks representing different road types. These blocks are straight, circular, ramps, roundabout, intersection, T-intersection, and forks. Because of the procedural road generation, this simulator is excellent for preventing overfitting to maps the software is trained on, making it a powerful and unique simulator among the other options. Fig. \ref{fig:PGDriveFig} shows screenshot of PGDrive map and graphics in simulation.
   
    \subsection{SUMO}
    SUMO \cite{c35} is a traffic simulator that simulates the flow of traffic and is used to simulate vehicle-to-vehicle interactions and path planning. Unlike the other simulators, it does not provide any sensors or graphics; it simply represents roads as graphs and simulates the paths taken by vehicles on the road. Thus, SUMO focuses on the problems of optimizing paths and traffic flow among agents on the road.

    \subsection{Flow}
    Flow \cite{c36} is a simulator built on top of SUMO; therefore, it also deals with the problem of traffic flow, path planning, and vehicle-to-vehicle communication. Flow was designed to simulate mixed-traffic cases, in which traffic has a large number of both autonomous and human-controlled vehicles. Flow features a modular design to easily create various scenarios to test and train autonomous vehicle models. Modules that can be controlled include the road layout, type and behavior of actors, and initial conditions. Maps can also be input from Open Street Map. 
    
\section{Embedded Operating Systems}
\label{sec:EmbeddedOS}
Embedded operating systems run autonomous vehicle software. They have the advantage of directly interfacing with the hardware on board, and the lack of software between the operating system and the hardware helps improve the system's security. Furthermore, many of the embedded operating systems used in autonomous vehicles are real-time operating systems (RTOS). Unlike traditional operating systems, they do not have a buffer and may include a prioritized scheduler to run critical actions first. The utility of the RTOS is that in an unexpected event that the vehicle must respond to, the operating system can immediately execute commands. Many embedded operating systems can also leverage a hypervisor, which allows the embedded operating system to run any number of additional operating systems or containers on top of it. The embedded operating system also isolates and contains the added operating systems to make them secure \cite{c48}. The isolation prevents infotainment systems or other software from being exploited. For example, infotainment operating systems like Android Auto or Apple CarPlay can operate securely on the hypervisor.
\medskip

Embedded operating systems used for autonomous vehicles must also be compatible with the Robot Operating System (ROS), an open-sourced set of tools and libraries for robot development \cite{c63}. It provides a platform to control process nodes and internal software communication within a robot.  Additionally, packages and libraries useful for autonomous vehicle development are available as ROS packages, such as the SLAM algorithms mentioned above.

\medskip

Security is of the highest importance in an autonomous vehicle, and because of this, formal security standards must be adhered to. A widely used certification for autonomous vehicles is ISO 26262 \cite{c73}. This standard defines four levels of Automotive Safety Integrity Levels (ASIL). There are levels A through D, with D being the highest level of security. ASIL D applies to systems with a high probability of harm or danger to life, and ASIL D certifications deem systems prone to such events capable and secure enough to handle them.

\medskip

There are many embedded operating systems on the market which provide platforms for the development of ADAS. Only one of these platforms is an open-source system, with the rest being proprietary products. The open-source embedded operating system is Automotive Grade Linux (AGL) \cite{c44}, an embedded operating system for autonomous vehicles. Unlike its competitors, it is not an RTOS, nor is it certified under ISO 26262. It is compatible with open-source hypervisors such as XEN \cite{XEN}, or Jailhouse \cite{Jailhouse}. AGL addresses all software components of a vehicle, such as infotainment, HUD, instrument cluster, and ADAS. Proprietary options for embedded operating systems are QNX Neutrino \cite{c39}, WindRiver VxWorks \cite{c40}, GreenHills INTEGRITY \cite{c41}, Nucleus OS \cite{c43}, and NVIDIA Drive \cite{c42}. All these operating systems come certified with ASIL D level of security certification. They are all RTOSs, have the option of using a hypervisor, and are compatible with ROS. Some RTOSs are already in use in vehicles, with 175 million vehicles worldwide using QNX Neutrino \cite{c39} and 36 thousand level 3-capable autonomous vehicles using WindRiver's VxWorks  \cite{c40}.

\medskip

The embedded operating system is a crucial consideration when designing secure and real-time software. ISO 26262 is a valuable security standard that can certify the operating system for autonomous vehicle applications. We discussed the value of RTOS technology in achieving real-time vehicle response. Finally, we highlighted the utility of hypervisors in creating software systems that deal with operating systems outside of the autonomous driving software. AGL is a free and open option for the embedded operating system, but it lacks ISO certification and is not an RTOS. Those developing an autonomous vehicle must evaluate these details when designing their vehicle. 

\section{Vehicle-to-Vehicle Technology}
\label{sec:V2X}
Vehicle-to-Everything (V2X) and Internet-of-Vehicle (IoV) are two promising fields relevant to autonomous vehicles. They center around connecting vehicles, infrastructure, pedestrians, and the cloud together. The aim is to leverage the information from all the participants in the network to enhance ADAS, optimize traffic flow, and provide infotainment to vehicle passengers.

    \subsection{V2X}
    \begin{figure*}[h]
        \centering
        \includegraphics[width=0.8\textwidth]{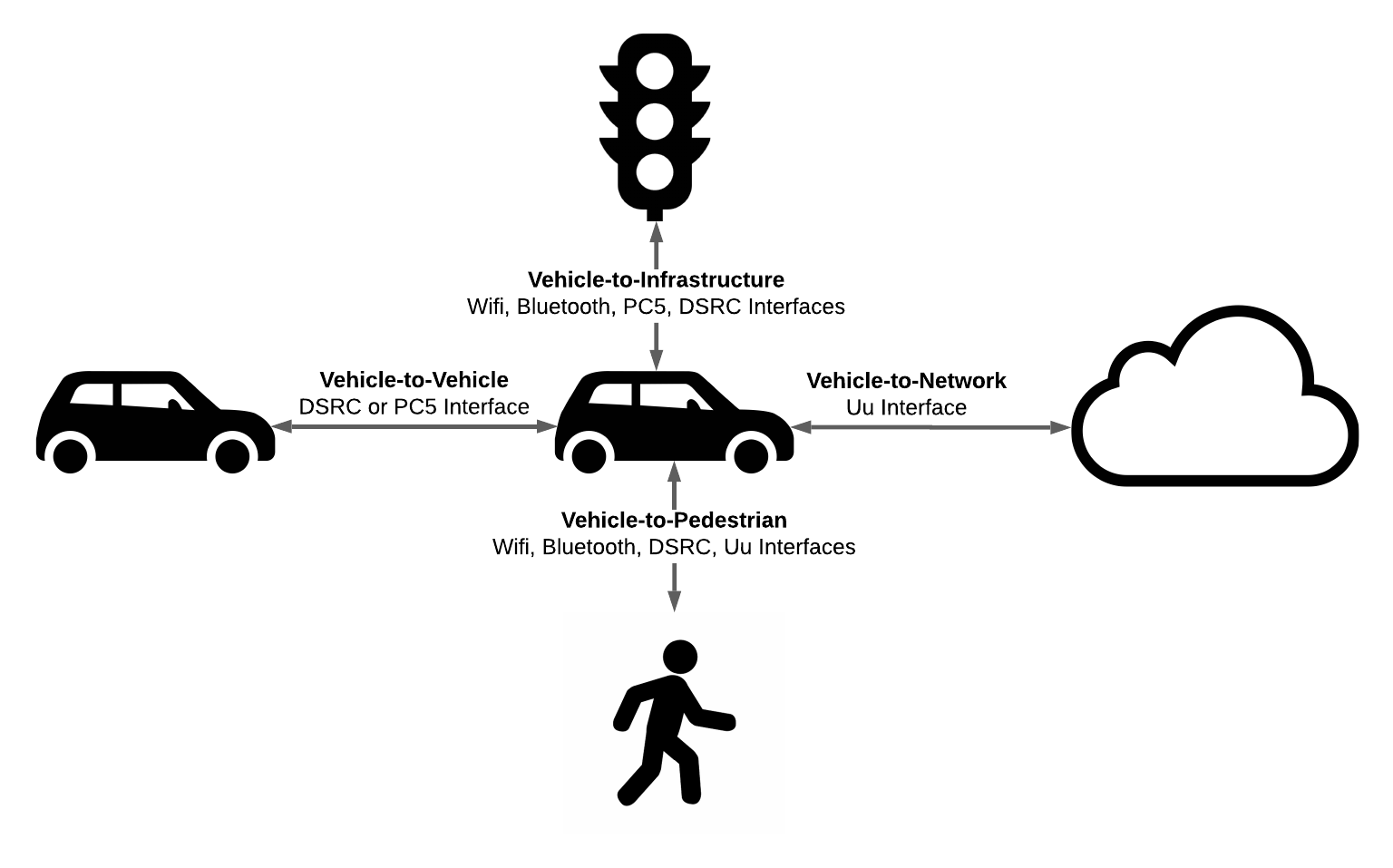}
        \caption{V2X Network and the possible networking interfaces}
        \label{fig:NetworkFig}
    \end{figure*}
    V2X relays traffic information between vehicles, infrastructure, pedestrians, and the cloud. It encompasses the following classes of communication: Vehicle-to-Vehicle (V2V), Vehicle-to-Pedestrian (V2P), Vehicle-to-Infrastructure (V2I), and Vehicle-to-Network (V2N) \cite{V2XTech}, \cite{c74}, \cite{c75}. V2V is direct communication between vehicles. The data transferred would help relay sensor, mapping, and safety information. It can improve the safety of ADAS by sharing blind spot information and coordinating vehicle actions \cite{V2V}. V2P allows bicyclists, pedestrians, and two-wheeled motor vehicles to relay location information so that vehicles can more easily avoid these road users \cite{V2P}. V2I connects to Road-side Units (RSU) and traffic lights to the network. V2I helps to connect infrastructure and vehicles to offload computations in the vehicles or facilitate large-scale traffic optimization. V2N connects vehicles to broader networks so that they may leverage cloud services or allow for large-scale optimizations. One of the primary goals is to improve driving safety by relaying location or SLAM information between cars to share information external to a vehicle's line of sight. \cite{ClassicalSLAM} discusses how V2X can be used for multi-vehicle SLAM, creating a unified centralized or decentralized digital map of the environment. The centralized model would utilize V2N, whereas a decentralized method would rely on V2V and V2I. This information can help the car make an appropriate autonomous decision to avoid collision in cases where the vehicle or even the driver would never have been able to. The NHTSA reports that at least 41\% of intersection collisions could be avoided with V2X technology and prevent a maximum of 400,000-500,000 crashes annually in just the US \cite{V2V}. In addition to its safety improvement, V2X can optimize traffic flow by coordinating vehicles using one or a combination of V2I, V2V, V2N. V2X can lead to smoother traffic and shorter wait times. Furthermore, V2X communication technology promises to reduce the environmental impact of traffic by optimizing its flow to minimize emissions \cite{V2XTech}. Fig. \ref{fig:NetworkFig} shows a schematic view of V2X network and the possible networking interfaces.
    
    \subsection{IoV}
    IoV is an extension of V2X technology that seeks to integrate cloud and artificial intelligence (AI) services into the autonomous vehicle ecosystem \cite{V2XTech}. It leverages big data analytics to optimize the network connectivity, allow for the expansion of infotainment options into augmented reality (AR), virtual reality (VR), and streaming, and provide novel services such as remote chauffeuring or care sharing \cite{c81}, \cite{V2XTech}. Many solutions have been proposed to extend the power of IoV through edge computing \cite{c67}. Edge computing can assist in offloading computation to edge devices or as caching for the network, and network devices \cite{AIECCaching}.
    
    \subsection{Network Interfaces}
    V2X technology typically uses the following networking technologies: Dedicated Short-Range Communication (DSRC), Cellular V2X (C-V2X) PC5 and Uu interfaces, wifi, and Bluetooth. They can be used in tandem to create a heterogeneous network \cite{DSRCCV2V} or individually. DSRC typically covers V2V and V2I classes of communication. It utilizes limited-range radio waves to send messages between vehicles and infrastructure on the road. The range permits communication up to 1000 meters and can transmit small amounts of data \cite{c81}, \cite{V2XTech}. DSRC can transmit SLAM information to enhance ADAS security and provide information about areas out of a car's direct line of sight. This unique advantage can improve the security of driverless vehicles beyond the level humans can achieve. Vehicles can be shipped with OEM DSRC devices or retrofitted with aftermarket DSRC devices, enabling quick and widespread adoption \cite{V2V}. Because DSRC must relay safety-critical information with low latency, it must adhere to the security standard IEEE 802.11p. 
    
    The second category of interface commonly used is C-V2X. It encompasses the V2N, V2I, and V2V classes and enables communication with other vehicles, cloud servers, and infrastructure. It leverages both the existing LTE network and upcoming 5G communication platforms. The introduction of 5G will enable low latency communication, pushing the expected capabilities of C-V2X to allow functions in levels 3 and 4 of automation \cite{5G_CV2X}. C-V2X accomplishes this with two different interfaces that use existing LTE technology: PC5 and Uu \cite{c74}. The PC5 interface can operate in and out of coverage for direct V2V communication. It has a 100m range and a low latency, making it suitable for latency-sensitive applications \cite{c75}. It allows direct communication between vehicles or infrastructure without the need to traverse the cellular network. The second interface for C-V2X is Uu technology, which transmits messages from a server through LTE infrastructure to vehicles. It has a higher latency, so it is better suited for latency tolerant cases like coordinating traffic or streaming infotainment \cite{c75}. It requires the vehicle to be within the coverage of the LTE network.

    Wifi and Bluetooth are network interfaces used to transmit data within the vehicular network. They have been proposed for V2P \cite{V2P}, and V2I \cite{V2I} applications. However, both of them face challenges. Wifi has latency issues in scenarios where vehicles can be moving up to 100 km/h \cite{V2P}. Bluetooth has a very short range of 50m, allowing it to work in some collision-prevention cases, but it may not be sufficient to carry out V2P communication fully.
    
    \subsection{Network Noise}
    A significant hurdle to both C-V2X and DSRC is network noise \cite{c78}. C-V2X suffers from high network load in dense traffic, exacerbated by pedestrian or other use of the LTE infrastructure. Similarly, too much load on the DSRC network can create a broadcast storm, causing package collision and slow down the network. Several solutions have been proposed to the problem of broadcast storms \cite{c77}, \cite{c79}, \cite{c80}, \cite{c80.5}. Despite the options, there is no standard adopted yet. 

    \subsection{Security}
    Network security is a vital issue to consider when using V2V and IoV technology. \cite{SecuringV2X}, \cite{V2VSecurity}, and \cite{ThreatToV2V} detail the different types of attacks on the vehicle network and system. \cite{SecuringV2X} categorizes network attacks into denial of service (DOS), Sybil attacks, and false data injection. DOS attacks may involve delaying information packet propagation or dropping it entirely. Sybil attacks involve an adversary taking on multiple identities to falsely show road congestion or increase vehicle reputation or trust score within the network. False data injections can use fabricated messages or relay an old message as a new one. \cite{ThreatToV2V} focuses on threats to DSRC and details solutions to mitigate some of the issues. Examples of threats include denial of service by flooding nodes with messages and jamming signals, spam causing transmission latency, injecting false messages into the network, or black holes that fail to propagate messages through the network. \cite{PrivacyTrustSecurity} divides the problems in vehicular networks into security, privacy, and trust.  Security is the problem of making network communication channels secure against adversarial attacks. Privacy deals with controlling the access of controlling vehicular information such as location and identity to the proper entities. Trust must be established between vehicles and with the messages they receive. \cite{NetworkSecuritySolutions} outlines many proposed solutions to each of the problems. Blockchain technology has also been proposed as a novel method of maintaining decentralized trust and privacy \cite{Blockchain}.
    
\section{Advanced Driving Assistance Systems}
\label{sec:ADAS}
Advanced Driving Assistance Systems (ADAS) facilitates the partial or complete automation of driving activities. Several levels of automation categorize ADAS. This section will introduce the five levels of automation as defined by the SAE \cite{SAE}. Following the introduction of the categories, some of the commercially available ADAS will be highlighted and discussed.
    \subsection{Five Levels of Automation}
   The five levels of automation classify autonomous vehicle systems from level 0, with zero automation, to fully autonomous driving with level 5. Each level has precise specifications, and the features that a vehicle possesses would place it within these levels. We present the definitions and examples in Table \ref{table:Levels}.

\label{table:Levels}
    \begin{table*}[t]
        \resizebox{\linewidth }{!} {
            \begin{tabular}{ | p{0.20\linewidth} ||  p{0.45\linewidth}| p{0.35\linewidth} | }
                \hline
                Level of Autonomoy& Definition & Example features\\
                \hline
                Level 0 & No driving automation. &Cruise control and ABS are considered non-autonomous functionalities.\\
                \hline
                Level 1 (Driver Assistance) & Execute either longitudinal or lateral vehicle control. Longitudinal includes acceleration and breaking, while lateral control involves steering. Driver must always be present and will perform all other driving tasks. & Active cruise control and lane departure warning system.\cite{SAELevelExamples}\\
                \hline
                Level 2 (Partial Driving Automation Level) & Vehicle automates a part of both lateral and longitudinal, but not all driving operations. The driver must be present to intervene if necessary. & The features at this level include lane keeping assist and parking assist.\\
                \hline
                Level 3 (Conditional Driving Automation) & Autonomous system performs the entire driving operations when engaged. System may request driver to intervene. & Features at this level include automatic emergency braking and active driving assistance.\\
                \hline
                Level 4 (High Driving Automation) & The system controls the entire driving operation while also handling the control fallback. This is in contrast to lower levels that have humans as the driving fallback. & Full autonomous driving subject to a condition, typically geo-fencing. Robotaxi fleets within a city would be an example of level 4 autonomy.\\
                \hline
                Level 5 (Full Driving Automation)& Full, unconditional vehicle autonomy. Driver intervention is unneeded as a fallback. & N/A\\
                \hline
            \end{tabular} 
        }\\
        \caption{A table summarizing features at each level of vehicle autonomy.}
    \end{table*}

    \subsection{Autonomous Vehicles in Industry}
    There are several proprietary and open-source ADAS available. The ADAS can be categorized using the five levels of vehicle automation depending on the features they bring to the table. NVIDIA Drive, Baidu Apollo, Autoware, and OpenPilot are assets to developers hoping to adapt existing open-source ADAS technology to their vehicles. Tesla is an example of a proprietary ADAS that is integrated into a manufacturer's lineup of cars. We present the above ADAS and their features in \ref{table:ADASTable}.

\label{table:ADASTable}
    \begin{table*}[t]
        \resizebox{\linewidth }{!} {
            \begin{tabular}{ | p{0.15\linewidth} || p{0.45\linewidth} | p{0.35\linewidth} | p{0.10\linewidth} | }
                \hline
                ADAS& Features & Hardware & Level Of Autonomy\\
                \hline
                NVIDIA Drive AGX \cite{c57}&Development kit with modules for image, point cloud, computer vision, obstacle perception, and path perception.& Camera and either Lidar or Radar&N/A \\
                \hline
                NVIDIA Drive Hyperion \cite{c57} &SLAM, Lane Planning, Active Cruise Control, Lane Keeping, Lane Merging, traffic light detection, intersection detection, semantic segmentation, obstacle detection, and path detection.&8 cameras and 8 radars&Level 2 and up \\
                \hline
                Baidu Apollo\cite{c58} &SLAM, perception, prediction, path planning, VX communication. &None&Level 4\\
                \hline
                Open Pilot \cite{c59} & Active cruise control, automatic lane centering, lane detection warning, forward collision warning.& Hardware kit with forward-facing camera. & Level 2\\
                \hline
                Autoware \cite{c59.5, Autoware}& Scene recognition, path planning, vehicle control for valet parking and cargo delivery. &None& N/A\\
                \hline
                Tesla \cite{c60} & Active cruise control, auto-steer, automatic lane change, automatic parking, sumonning, traffic and stop sign control, automatic emergency braking, front and side collision warning, obstacle aware acceleration, blind spot monitoring, lane steering. & 8 cameras, a radar, and 12 ultrasonic sensors. & Level 2\\
                \hline
            \end{tabular} 
        }\\
        \caption{A table summarizing features and sensors available in Advanced Driving Assistance Systems}
    \end{table*}

\section{Commercial Autonomous Vehicle Companies}
\label{sec:Commercial}
Traditional automotive manufacturers have partnered with and acquired software companies to integrate autonomous vehicle technology into their lineups. Partnerships allow companies to share expertise and work towards a common goal. We will outline highlights from \cite{CBInsights} with a section about manufacturers and a section about software companies. The first sections below will discuss the prominent manufacturers, their acquisitions, partnerships, and delivery dates. The second section will outline companies that provide software solutions for autonomous vehicles, testing, and results.

    \subsection{Manufacturers}
    Car manufacturers have been utilizing partnerships and acquisitions to fill the knowledge and skill gaps with the technology in autonomous vehicles. Companies have bolstered their expertise with these acquisitions, and some have successfully tested vehicles on the streets. With the positive results of the testing, many manufacturers have set their expected deployment dates to the near future. Tesla, one of the first to bring autonomous vehicle technology to the market, has had autonomous driving features in its vehicles since 2014. In 2019 they added the option for hardware kits that expand their vehicles to level 3 autonomous driving. BMW, Intel, and Mobileye partnered to develop autonomous vehicles and plan to deploy fleets in 2021. Argo AI, a prominent autonomous vehicle startup, was acquired by Ford and is partnered with Audi to deliver its driverless technology. With Argo AI, Ford aims to begin rolling out autonomous vehicles starting in 2022. GM and Honda have partnered together to develop their technologies together. Honda plans to mass produce level 3 autonomous vehicles. GM has acquired Cruise Automation and has launched semi-autonomous driving capabilities in its vehicles since 2018. FAW Group launched a robotaxi service in 2019, partnered with Baidu. Bosch has also launched a trial for robotaxis in California and has invested \$1 billion in manufacturing semiconductors to use smart cities and autonomous vehicles. Though Tesla and GM have led the way for autonomous vehicle features, the coming years will see many more autonomous vehicles commercially available and on the roads. 
    
    \subsection{Software Companies}
    The most challenging part of developing an autonomous vehicle is the complex software that perceives its environment, plans its motion, and executes its trajectory. Many software giants and brand-new startups have worked to tackle the challenge. Major players such as Amazon and Apple have heavily invested in the field by acquiring companies and talent, though there are no commercial offerings or announcements from either of them. Alphabet's Waymo has been testing and developing autonomous vehicles for many years. They have achieved an industry-leading ten billion miles of simulated driving and twenty million miles of real-world driving. They have been providing self-driving services since 2018 with a sixty-two thousand car fleet. They are partnered with UPS as well as Fiat-Chrysler to bring their technologies to the road. The strongest player in the Chinese market, Baidu, has similarly impressive technology with their Apollo open-source operating system and self-driving platform. Over 150 partners worldwide use the platform, including Chevrolet, Ford, Honda, Toyota, and Volkswagen. They boast 3 million kilometers of autonomous vehicle road training. Another promising candidate for autonomous vehicle software is Aptiv. They have already deployed robotaxi operations in partnership with Lyft to provide service in Las Vegas, Boston, Singapore, and Pittsburgh. Aptiv also has a joint venture with Hyundai called Motional. Motional aims to commercialize robotaxi fleets starting in 2022. Along with Baidu, Aptiv aims to capitalize on the Chinese market, which is expected to account for more than two-thirds of the autonomous vehicle market in the world.

\section{Conclusion}
Autonomous vehicles are complex software, networking, and hardware systems that can leverage open-source code to rapidly and cost-effectively develop. We have outlined key design considerations and open-source resources for designing autonomous vehicles such as SLAM libraries, open-source simulators, embedded operating systems, V2X, IoV, and ADAS platforms. We have also introduced the reader to the levels of vehicle automation according to the SAE and given an overview of the autonomous vehicle market manufacturers and software companies. The broad understanding of the field and the open-source tools to start equips a developer to begin work on their own autonomous vehicle solutions.

\section*{Acknowledgement}

This project is supported by Future Skills Centre, Canada

\ifCLASSOPTIONcaptionsoff
  \newpage
\fi



\bibliographystyle{IEEEtran}
\bibliography{AVT.bib}
%

%
\begin{IEEEbiography}[{\includegraphics[width=1in,height=1.25in,clip,keepaspectratio]{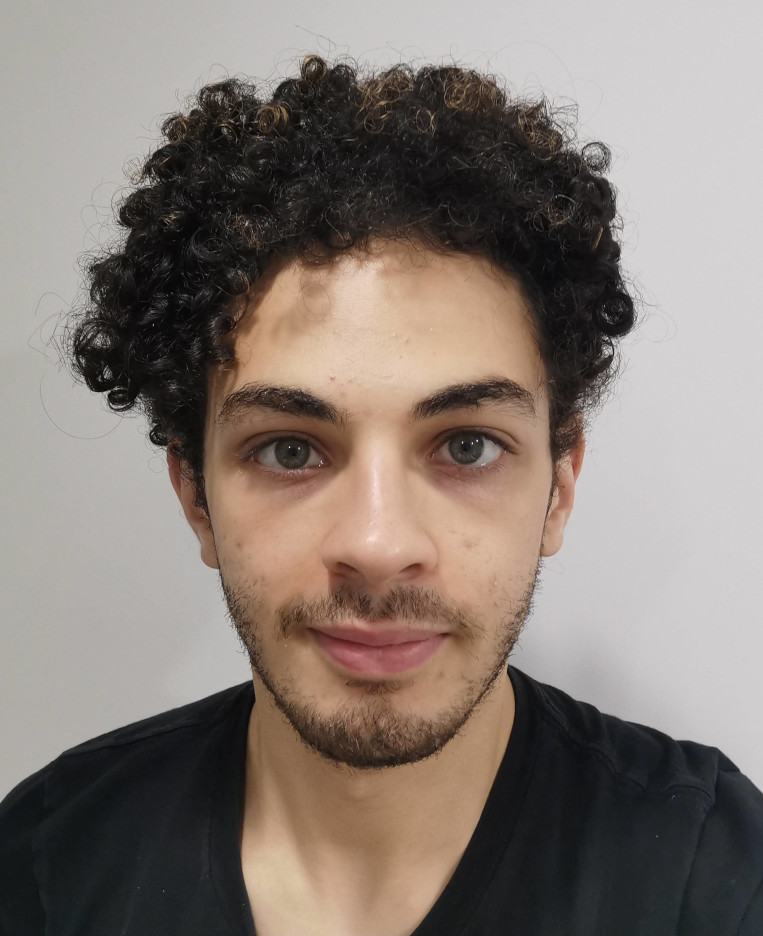}}]{Oussama Saoudi}
Oussama Saoudi is an undergraduate student in the Department of Computing and Software at McMaster University. His research interests lie in artificial intelligence, robotics, and autonomous vehicles.
\end{IEEEbiography}
\begin{IEEEbiography}[{\includegraphics[width=1in,height=1.25in,clip,keepaspectratio]{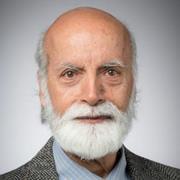}}]{Ishwar Singh} 
is an Adjunct Professor in the School of Engineering Practice and Technology (SEPT) at McMaster University since 2011. Prior to that he was the founding Associate Director of the four-year BTech programs, a joint venture between McMaster University and Mohawk College, and the Chair of the process automation program. He coordinated the curriculum design development and implementation of the process automation, automotive and vehicle technology, and biotechnology programs in addition to his contribution for the establishment of the energy engineering technology degree completion B.Tech. program. More recently he has been involved in the establishment of the SEPT Learning Factory for Industry 4.0 Education, Training and Applied Research, and M.Eng. programs.
\end{IEEEbiography}
\begin{IEEEbiography}[{\includegraphics[width=1in,height=1.25in,clip,keepaspectratio]{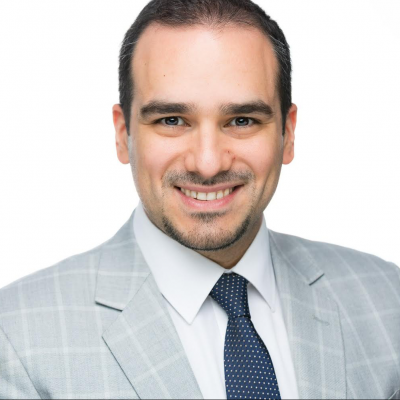}}]{Hamidreza Mahyar} 
is an Assistant Professor in the School of Engineering Practice and Technology and an Associate Member in the Department of Computing and Software in the Faculty of Engineering at McMaster University. Before joining MacMaster, he was a postdoctoral research fellow at Boston University and Technical University of Vienna, working with Prof. Eugene Stanley. Dr. Mahyar received his Ph.D. in Computer Engineering from Sharif University of Technology. His research interests lie at the intersection of machine learning, network science, and natural language processing, with a current emphasis on the fast-growing subject of graph neural networks with applications in recommendation systems, drug discovery, and self-driving cars.
\end{IEEEbiography}




\end{document}